\documentclass[conference]{IEEEtran}

\IEEEoverridecommandlockouts
\usepackage{cite}
\usepackage{amsmath,amssymb,amsfonts}
\usepackage{algorithm} 
\usepackage{algpseudocode}
\usepackage{float}
\usepackage{graphicx}
\usepackage{textcomp}
\usepackage{xcolor}
\usepackage{url}
\usepackage{booktabs}
\usepackage{multirow}
\usepackage{threeparttable}
\usepackage[letterpaper, left=0.65in, right=0.65in, top=0.75in, bottom=1.2in]{geometry}

\def\BibTeX{{\rm B\kern-.05em{\sc i\kern-.025em b}\kern-.08em
    T\kern-.1667em\lower.7ex\hbox{E}\kern-.125emX}}
\begin{document}

\title{L2M-AID: Autonomous Cyber-Physical Defense by Fusing Semantic Reasoning of Large Language Models with Multi-Agent Reinforcement Learning 
}

\author{
Tianxiang Xu\textsuperscript{1$\dagger$}\thanks{$\dagger$~These authors contributed equally to this work},
Zhichao Wen\textsuperscript{2$\dagger$},
Xinyu Zhao \textsuperscript{3},
Jun Wang\textsuperscript{4},
Yan Li\textsuperscript{5},
Chang Liu\textsuperscript{6*}\thanks{*~Corresponding author: Chang Liu (cryyu1478@outlook.com)}
\\[0.5ex]
\textsuperscript{1}Peking University, Beijing, China\\
\textsuperscript{2}RWTH Aachen University, Aachen, Germany\\
\textsuperscript{3}University of Texas at Austin, Austin, USA\\
\textsuperscript{4}Wuhan University, Wuhan, China\\
\textsuperscript{5}Thales Group, Ottawa, Canada\\
\textsuperscript{6}Chinese Medical Information and Big Data Association, Beijing, China
}

\maketitle

\footnotetext{\small For any commercial use or derivative works, please contact the IEEE Copyrights Office at copyrights@ieee.org.}

\begin{abstract}
The increasing integration of Industrial IoT (IIoT) exposes critical cyber-physical systems to sophisticated, multi-stage attacks that elude traditional defenses lacking contextual awareness. This paper introduces L2M-AID, a novel framework for Autonomous Industrial Defense using LLM-empowered, Multi-agent reinforcement learning. L2M-AID orchestrates a team of collaborative agents, each driven by a Large Language Model (LLM), to achieve adaptive and resilient security. The core innovation lies in the deep fusion of two AI paradigms: we leverage an LLM as a semantic bridge to translate vast, unstructured telemetry into a rich, contextual state representation, enabling agents to reason about adversary intent rather than merely matching patterns. This semantically-aware state empowers a Multi-Agent Reinforcement Learning (MARL) algorithm, MAPPO, to learn complex cooperative strategies. The MARL reward function is uniquely engineered to balance security objectives (threat neutralization) with operational imperatives, explicitly penalizing actions that disrupt physical process stability. To validate our approach, we conduct extensive experiments on the benchmark SWaT dataset and a novel synthetic dataset generated based on the MITRE ATT\&CK for ICS framework. Results demonstrate that L2M-AID significantly outperforms traditional IDS, deep learning anomaly detectors, and single-agent RL baselines across key metrics, achieving a 97.2\% detection rate while reducing false positives by over 80\% and improving response times by a factor of four. Crucially, it demonstrates superior performance in maintaining physical process stability, presenting a robust new paradigm for securing critical national infrastructure.
\end{abstract}

\begin{IEEEkeywords}
cyber-physical systems security, multi-agent reinforcement learning, large language models, autonomous defense, semantic reasoning
\end{IEEEkeywords}

\section{Introduction}

The fourth industrial revolution is catalyzing a profound convergence of Operational Technology (OT) and Information Technology (IT), forging hyper-connected Industrial IoT (IIoT) ecosystems that promise unprecedented efficiency gains but simultaneously introduce a perilous new attack surface \cite{xu2018industry}. The dissolution of the traditional air-gap has interwoven once-isolated industrial control systems (ICS) with enterprise networks, exposing critical infrastructure to sophisticated cyber-physical threats where digital intrusions inflict tangible physical damage \cite{ferrag2023deep}. Seminal events like the Stuxnet worm's sabotage of nuclear centrifuges \cite{langner2011stuxnet} and the Mirai botnet's weaponization of IoT devices \cite{antonakakis2017mirai} are not mere historical footnotes, but harbingers of a reality where securing these deeply coupled systems demands a fundamental rethinking of our defense posture. Confronted with this landscape, existing security paradigms are proving critically inadequate. Signature-based Intrusion Detection Systems (SIDS) are inherently blind to the zero-day exploits that characterize modern APTs \cite{khraisat2019survey}, while Anomaly-based Intrusion Detection Systems (AIDS) face a fundamental paradox in IIoT. The very predictability of industrial processes, which simplifies baseline modeling, is now being exploited by adversaries to orchestrate ``low-and-slow'' attacks that mimic legitimate traffic to evade detection \cite{gusmao2018survey, giraldo2018survey}. This underscores a critical flaw: current systems are sophisticated pattern recognizers but lack genuine intent understanding. They can flag a statistical anomaly but cannot discern the malicious intent behind a sequence of seemingly benign actions, a deficiency exacerbated by the operational constraints of resource-limited and often unpatchable legacy systems \cite{al2019survey}.

To transcend these limitations, we advocate for a paradigm shift from passive detection to proactive, autonomous defense, driven by breakthroughs in artificial intelligence. The advent of Large Language Models (LLMs) offers a transformative capability, evolving them from mere language processors into powerful reasoning engines capable of synthesizing vast, heterogeneous data streams---from cryptic system logs to open-source threat intelligence---into a coherent, causal understanding of an unfolding security event \cite{he2024survey, wu2023autogen}. However, in the distributed fabric of IIoT, a monolithic intelligence is a bottleneck and a single point of failure. This necessitates a Multi-Agent System (MAS) architecture, where a decentralized team of specialized agents can offer scalable and resilient defense \cite{ferguson2025creating}. This paper introduces L2M-AID, a framework that materializes this vision. Our central thesis is that a deep fusion of the high-level semantic reasoning of LLMs with the low-level, adaptive control of Multi-Agent Reinforcement Learning (MARL) can bridge the chasm between understanding and action. L2M-AID orchestrates a hierarchical team of LLM-empowered agents, trained collectively via the MAPPO algorithm under a Centralized Training, Decentralized Execution (CTDE) paradigm \cite{lowe2017multi}. Crucially, our MARL formulation is tailored for the cyber-physical domain, with a reward function meticulously engineered to balance the dual objectives of threat neutralization and the preservation of physical process stability.

The primary contributions of this work are threefold:
\begin{enumerate}
    \item \textbf{A Novel Hierarchical Multi-Agent Framework:} We design and propose L2M-AID, a role-based, hierarchical agent architecture that synergistically fuses LLM-driven semantic reasoning with MARL-based adaptive control, creating a novel solution for autonomous cyber-physical defense.
    \item \textbf{A Context-Aware MARL Formulation for IIoT:} We introduce a formal MARL model where the state space is semantically enriched by LLM-generated contextual embeddings, and the reward function explicitly co-optimizes for security efficacy and physical process stability, bridging the gap between cyber defense and operational safety.
    \item \textbf{Comprehensive Empirical Validation and Generalization:} We conduct rigorous experiments on the benchmark SWaT dataset and a novel synthetic attack dataset derived from the MITRE ATT\&CK for ICS framework, demonstrating that L2M-AID achieves superior performance and generalization against a spectrum of baselines.
\end{enumerate}

\section{Related Work}
This section critically reviews three core research domains integral to our work: intrusion detection in Industrial Control Systems (ICS), the application of Large Language Models in cybersecurity, and Multi-Agent Reinforcement Learning for network defense. By analyzing the trajectory and inherent limitations within each domain, we delineate the research chasm that motivates the novel architectural fusion presented in our L2M-AID framework.

\subsection{The Evolution and Stalemate in ICS Intrusion Detection}
Research in ICS intrusion detection has progressed from static techniques to sophisticated data-driven models, yet a fundamental gap in contextual understanding persists. The complexity of these environments necessitates validation on high-fidelity datasets sourced from specialized testbeds like SWaT \cite{mathur2016swat}. Early defense strategies centered on graphical model-based approaches for anomaly detection \cite{hadinoto2009anomaly}, which, while precise, lacked flexibility against novel attacks. The field then advanced towards machine learning, with hybrid algorithms combining techniques like vector quantization and one-class SVMs showing promise \cite{fawzy2018unsupervised}. This was followed by the rise of deep learning, which demonstrated strong capabilities in omni-directional SCADA intrusion detection \cite{maglaras2016combining} and specialized architectures like 1D-CNN autoencoders for specific cyber-physical tasks like leakage detection \cite{lee2018applying}.

Despite the increasing sophistication of machine learning applications in ICS security \cite{goh2017anomaly}, these models largely function as reactive pattern recognizers. Their core limitation is that they operate on raw or statistically-derived features, lacking the context to interpret the logical sequence of an attack. They can identify that a state is anomalous, but not \textit{why} it is malicious in the context of a broader campaign, such as those cataloged in the MITRE ATT\&CK for ICS framework \cite{mitre2020attack}. This semantic gap is the primary reason for their struggles with high false positive rates and their inability to counter attackers who adeptly mimic legitimate operational behavior.

\begin{figure}[t]
\centering
\includegraphics[width=\linewidth]{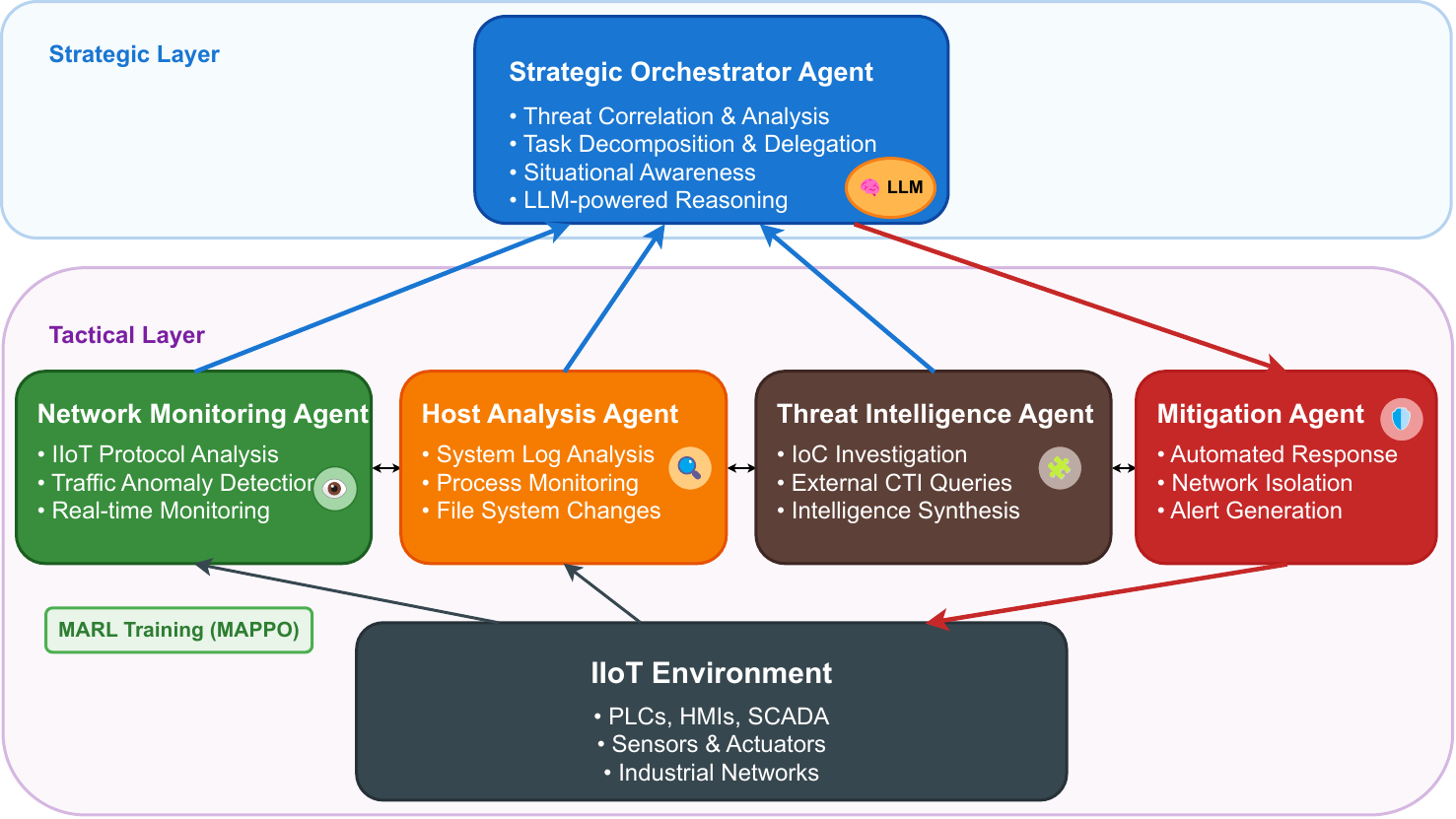} 
\caption{The hierarchical architecture of L2M-AID, illustrating the strategic Orchestrator Agent and the tactical Monitoring, Analysis, and Mitigation Agents. The solid lines represent the primary data and command flow, while the dashed lines indicate the broadcast of the LLM-generated contextual state embedding ($L_t$).}
\label{fig:architecture}
\end{figure}

\subsection{LLMs as Semantic Bridges for Cyber Threat Intelligence}
Large Language Models (LLMs) have emerged as a powerful solution to this semantic gap, introducing a paradigm of high-level reasoning to cybersecurity. Their primary application lies in automating the cognitive labor of human analysts, demonstrating profound capabilities in performing advanced malicious log analysis \cite{ferraro2023security} and extracting structured TTPs from unstructured threat intelligence \cite{he2024survey}. This has naturally led to their integration with SOAR platforms to automate incident response, where LLMs can triage alerts and suggest actions based on predefined playbooks \cite{das2024autonomous}. The latest evolution of this trend is the development of ``generative agents''---LLM-powered autonomous entities capable of complex, interactive behavior and planning \cite{park2023generative}.

\subsection{MARL for Adaptive Control and the Research Chasm}

Multi-Agent Reinforcement Learning (MARL) provides a robust framework for translating cognitive understanding into decentralized, adaptive control. Situated within the fully cooperative paradigm as outlined in recent reviews \cite{oroojlooy2023review}, MARL has gained prominence in network defense, with extensive studies exploring its applications in communication networks \cite{oliehoek2016concise}, including MARL-based intrusion detection systems \cite{zhang2021robust} and reinforcement learning for fine-grained tasks such as feature selection \cite{wang2020reinforcement}. The architectural evolution of multi-agent systems has been further advanced through formalization of AI agent communication protocols that ensure resilient coordination \cite{ibm2024acp}. These developments collectively enhance MARL’s stability and practicality, especially with algorithms like PPO proving effective in complex cooperative environments \cite{yu2022surprising}. Despite these advances, a fundamental gap persists between two powerful yet disconnected paradigms: LLMs excel in high-level symbolic reasoning but lack adaptive learning, while MARL offers dynamic sub-symbolic control yet struggles with semantically rich state spaces \cite{ng1999policy}. The core innovation of L2M-AID is bridging this divide through a neuro-symbolic fusion where the LLM acts as a dynamic ``state shaper'' and ``reward interpreter'' for the MARL algorithm. The LLM transforms raw, high-dimensional telemetry into semantically contextualized representations, enabling MARL agents to learn coordinated, context-aware strategies. In return, MARL provides the continuous adaptation that LLMs lack, forming a unified, intelligent defense framework capable of both reasoning and autonomous control.

\section{Methodology}

This section delineates the architectural design and theoretical underpinnings of the L2M-AID framework. We first present the hierarchical multi-agent architecture engineered for autonomous cyber defense. Subsequently, we provide a rigorous mathematical formalization of the defense problem as a cooperative multi-agent reinforcement learning task, detailing our novel state representation and reward structure. Finally, we describe the learning algorithm adopted to derive optimal collaborative defense policies.

\begin{figure*}[t]
\centering
\includegraphics[width=0.8\linewidth]{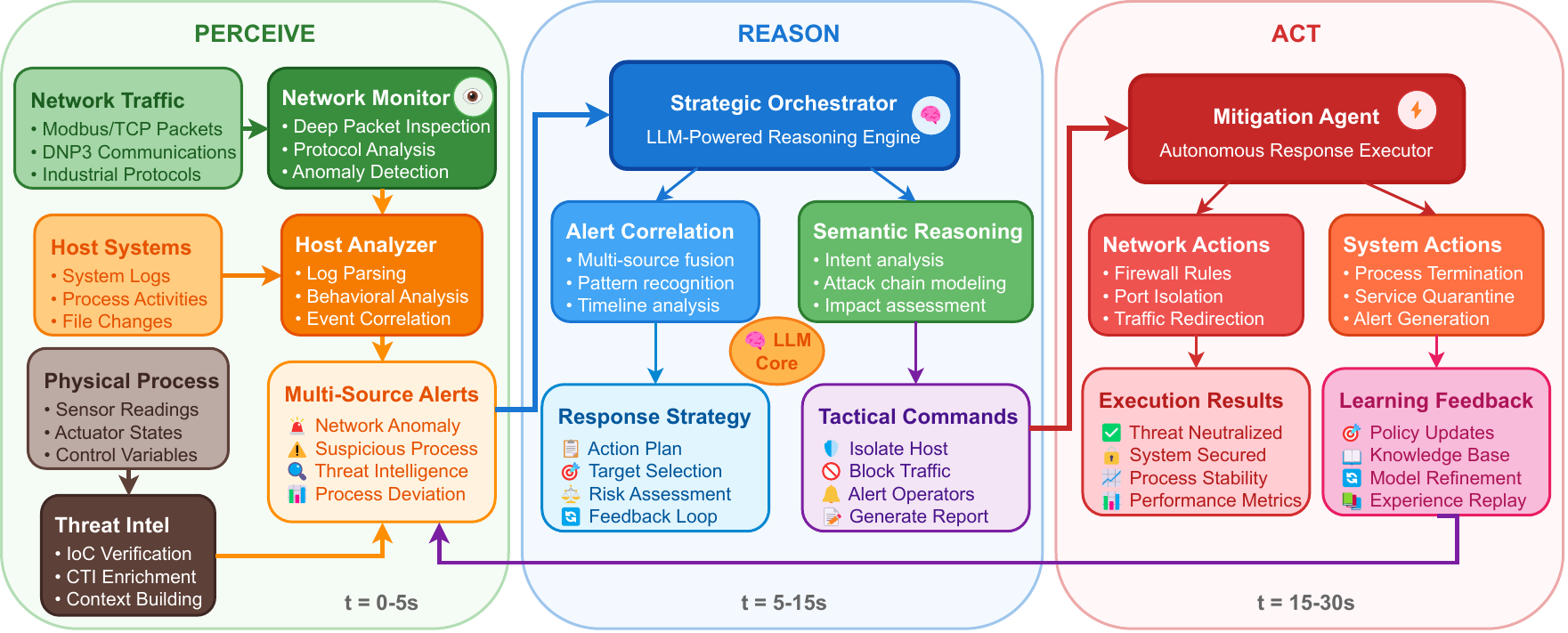} 
\caption{The operational data and decision pipeline of L2M-AID. Tactical agents convert raw data into alerts, which the Orchestrator correlates and reasons upon to formulate a strategy, finally commanding the Mitigation Agent to act.}
\label{fig:pipeline}
\end{figure*}

\subsection{Hierarchical Multi-Agent Architecture}

L2M-AID adopts a hierarchical, distributed multi-agent architecture inspired by the functional structure of a Security Operations Center (SOC). The system separates high-level reasoning from low-level execution through two layers: a \textbf{Strategic Orchestrator Agent} and a set of \textbf{Tactical Agents}. The Orchestrator, powered by a security-domain fine-tuned LLM, serves as the cognitive core—correlating multi-source alerts, assessing threats, planning strategic responses, and maintaining situational awareness through adaptive reasoning. The Tactical layer comprises specialized agents that sense, analyze, and act within the IIoT environment: the \textbf{Network Monitoring Agent} inspects industrial protocols and detects anomalies, the \textbf{Host Analysis Agent} interprets logs to uncover stealthy activities, the \textbf{Threat Intelligence Agent} enriches alerts via internal and external intelligence, and the \textbf{Mitigation Agent} executes safe, pre-authorized responses. Together, these agents form a continuous perceive–reason–act cycle that transforms raw telemetry into contextual understanding and coordinated defensive actions.

\subsection{Formalization as a Cooperative MARL Task}

We formalize the autonomous defense problem as a Decentralized Partially Observable Markov Decision Process (Dec-POMDP), defined by $\mathcal{M} = \langle \mathcal{S}, \mathcal{A}, P, \mathcal{R}, \mathcal{O}, \Omega, n, \gamma \rangle$, where each agent $i$ observes only $o_t^i \in \mathcal{O}$, a partial view of the global state $s_t \in \mathcal{S}$. This models real IIoT environments where centralized state observation is infeasible.

\subsubsection{State and Observation Space ($\mathcal{S}, \mathcal{O}$)}
The global state $s_t$ represents the complete IIoT environment at timestep $t$. Each agent $i$ perceives a multi-modal local observation:
\begin{equation}
\label{eq:state}
\mathbf{o}_t^i = \langle \mathcal{N}_t^i, \mathcal{H}_t^i, \mathcal{P}_t^i, \mathcal{C}_t^i, \mathcal{L}_t \rangle
\end{equation}
where $\mathcal{N}_t^i$ and $\mathcal{H}_t^i$ encode network and host telemetry; $\mathcal{P}_t^i$ contains critical physical process variables (pressure, temperature); and $\mathcal{C}_t^i$ buffers inter-agent messages.

Our key contribution is $\mathcal{L}_t$, a shared contextual embedding generated by the Orchestrator's LLM through periodic synthesis of aggregated alerts. This dense vector encodes high-level semantic concepts (e.g., "reconnaissance detected", "lateral movement in progress") and is broadcast to all agents, enriching local observations with global context. This mechanism provides implicit coordination, enabling agents to align behaviors despite partial observability—a fundamental Dec-POMDP challenge.

\subsubsection{Action Space ($\mathcal{A}$)}
The joint action space $\mathcal{A} = \times_{i=1}^n \mathcal{A}_i$ comprises discrete, finite action sets per agent. Each $\mathcal{A}_i$ is role-specific to ensure operational safety—e.g., the Mitigation Agent is restricted to reversible containment operations.

\subsubsection{Reward Function Engineering ($\mathcal{R}$)}
We engineer a shared global reward $\mathcal{R}(s, \mathbf{a})$ for joint action $\mathbf{a} = (a^1, \dots, a^n)$ that balances security efficacy with operational constraints:
\begin{equation}
\label{eq:reward}
\begin{split}
\mathcal{R}(s, \mathbf{a}) = \; & w_{sec} R_{\text{security}}(s, \mathbf{a}) 
+ w_{proc} R_{\text{process}}(s, \mathbf{a}) \\
& + w_{cost} R_{\text{cost}}(\mathbf{a})
\end{split}
\end{equation}
$R_{\text{security}}$ rewards threat neutralization and penalizes missed detections; $R_{\text{process}}$ enforces operational safety through penalties for deviations from safe bounds; $R_{\text{cost}}$ encourages efficiency via minor action and false positive penalties. The weights ($w_{sec}, w_{proc}, w_{cost}$) encode domain-specific risk tolerance, enabling customized defense policies.

\subsection{Learning Algorithm: MAPPO with Centralized Training}

To solve this Dec-POMDP, we employ Multi-Agent Proximal Policy Optimization (MAPPO), an algorithm that has demonstrated superior performance and stability in complex cooperative multi-agent settings \cite{yu2022surprising}. PPO's core mechanism, the clipped surrogate objective, prevents excessively large policy updates, which is crucial for stable learning in the non-stationary environment characteristic of multi-agent systems. The objective for each agent $i$'s policy $\pi_{\theta_i}$ is given by:
\begin{equation}
L(\theta_i) = \mathbb{E}_t \left[ \min(r_t(\theta_i) \hat{A}_t, \text{clip}(r_t(\theta_i), 1-\epsilon, 1+\epsilon) \hat{A}_t) \right]
\end{equation}
where $r_t(\theta_i) = \frac{\pi_{\theta_i}(a_t|o_t)}{\pi_{\theta_{i, \text{old}}}(a_t|o_t)}$ is the probability ratio and $\hat{A}_t$ is the estimated advantage function.

We adopt the Centralized Training for Decentralized Execution (CTDE) paradigm. During the centralized training phase, we introduce a centralized critic that has access to the global state $s$ (or the full set of observations $\{\mathbf{o}^1, \dots, \mathbf{o}^n\}$), enabling it to learn an accurate joint-action value function $V(s)$. This global perspective provides a stable and informative learning signal that effectively addresses the credit assignment problem and mitigates the non-stationarity arising from concurrently learning policies. For decentralized execution, once the training converges, the centralized critic is discarded. Each agent $i$ then operates autonomously in the deployment environment, executing its learned policy $\pi_{\theta_i}$ based solely on its local observation history. This decoupling ensures that the deployed system is scalable, robust, and maintains a low-latency response capability, as it eliminates the need for a centralized controller at runtime.

\section{Experimental Evaluation}

To rigorously assess the efficacy and cyber-physical safety of the L2M-AID framework, we conducted a series of comprehensive experiments. This section details the experimental setup, including the evaluation datasets and our novel synthetic attack generation methodology. We then specify the implementation details, the baseline models against which L2M-AID is benchmarked, and the metrics used for performance evaluation. Finally, we present and analyze the quantitative results, including a crucial ablation study, and provide a qualitative case study to illustrate the framework's operational intelligence.

\begin{table}[t]
\caption{Key Hyperparameter Configuration}
\label{tab:hyperparameters}
\centering
\scriptsize
\renewcommand{\arraystretch}{1.15} 
\begin{tabular}{p{2.1cm}|p{3.2cm}|p{2.3cm}}
\hline
\textbf{Category} & \textbf{Parameter} & \textbf{Value} \\
\hline
\textbf{MAPPO Algorithm} & Actor Learning Rate ($\alpha_{\text{actor}}$) & 5e-4 \\
 & Critic Learning Rate ($\alpha_{\text{critic}}$) & 5e-4 \\
 & Discount Factor ($\gamma$) & 0.99 \\
 & GAE Lambda ($\lambda$) & 0.95 \\
 & PPO Clip Parameter ($\epsilon$) & 0.2 \\
 & Training Epochs & 10 \\
\hline
\textbf{Neural Network} & Actor/Critic Network & 3-layer MLP \\
\textbf{Architecture} & Activation Function & ReLU \\
\hline
\textbf{LLM Parameters} & Base Model & Llama-3-8B-Instruct \\
 & Temperature & 0.2 \\
 & Top-p & 0.9 \\
\hline
\textbf{Reward Weights} & Security Weight ($w_{sec}$) & 1.0 \\
 & Process Stability Weight ($w_{proc}$) & 2.0 \\
 & Cost Weight ($w_{cost}$) & 0.1 \\
\hline
\end{tabular}
\end{table}


\subsection{Evaluation Datasets and Methodology}

Our evaluation leverages an offline replay methodology, where the defense framework processes datasets chronologically. At each timestep, agents make decisions based on their current observations, and the efficacy of these actions is evaluated against ground-truth labels. This approach enables a deterministic and reproducible assessment of the agents' learned policies.

\subsubsection{SWaT Benchmark Dataset}
The primary evaluation is performed on the Secure Water Treatment (SWaT) dataset, a widely recognized gold-standard benchmark for ICS security research \cite{mathur2016swat}. Generated from a fully operational, industrial-scale water treatment testbed, the dataset provides eleven days of continuous sensor and actuator data, synchronized with network traffic captures. This period includes seven days of normal operations and four days featuring 36 distinct multi-stage cyber-physical attacks, complete with precise temporal ground-truth labels. The high fidelity of SWaT and its integration of both cyber and physical data are crucial for validating L2M-AID's core competency in defending cyber-physical processes.

\subsubsection{Synthetic Attack Dataset with Conditional GANs}

To evaluate generalization against zero-day threats, we developed a synthetic dataset generation pipeline that transcends static benchmark limitations. Our four-stage methodology synthesizes physically-consistent attack scenarios derived from threat intelligence: 1) We establish benign operational baselines from seven days of normal SWaT data, extracting temporal patterns and cross-sensor correlations. 2) We formalize adversarial behavior as an attack graph based on MITRE ATT\&CK for ICS \cite{mitre2020attack}, where vertices represent tactics and weighted edges encode transition probabilities. 3) Treating this graph as an HMM, we perform probabilistic traversals to generate diverse TTP sequences absent from original SWaT attacks \cite{jeon2021probabilistic}. 4) We employ a conditional TimeGAN \cite{yoon2019time} to inject attacks while preserving physical consistency—the generator learns normal process dynamics and uses TTP sequences as conditioning inputs to guide anomaly injection. Crucially, our approach ensures physical coupling: when manifesting TTP T0831 "Manipulation of Control" on sensor `LIT101`, correlated sensors like `FIT101` exhibit consistent deviations based on hydraulic constraints. This physics-aware synthesis yields realistic, stealthy attacks that respect process dynamics, providing a rigorous generalization testbed.

\begin{figure}[t]
    \centering
    \includegraphics[width=\linewidth]{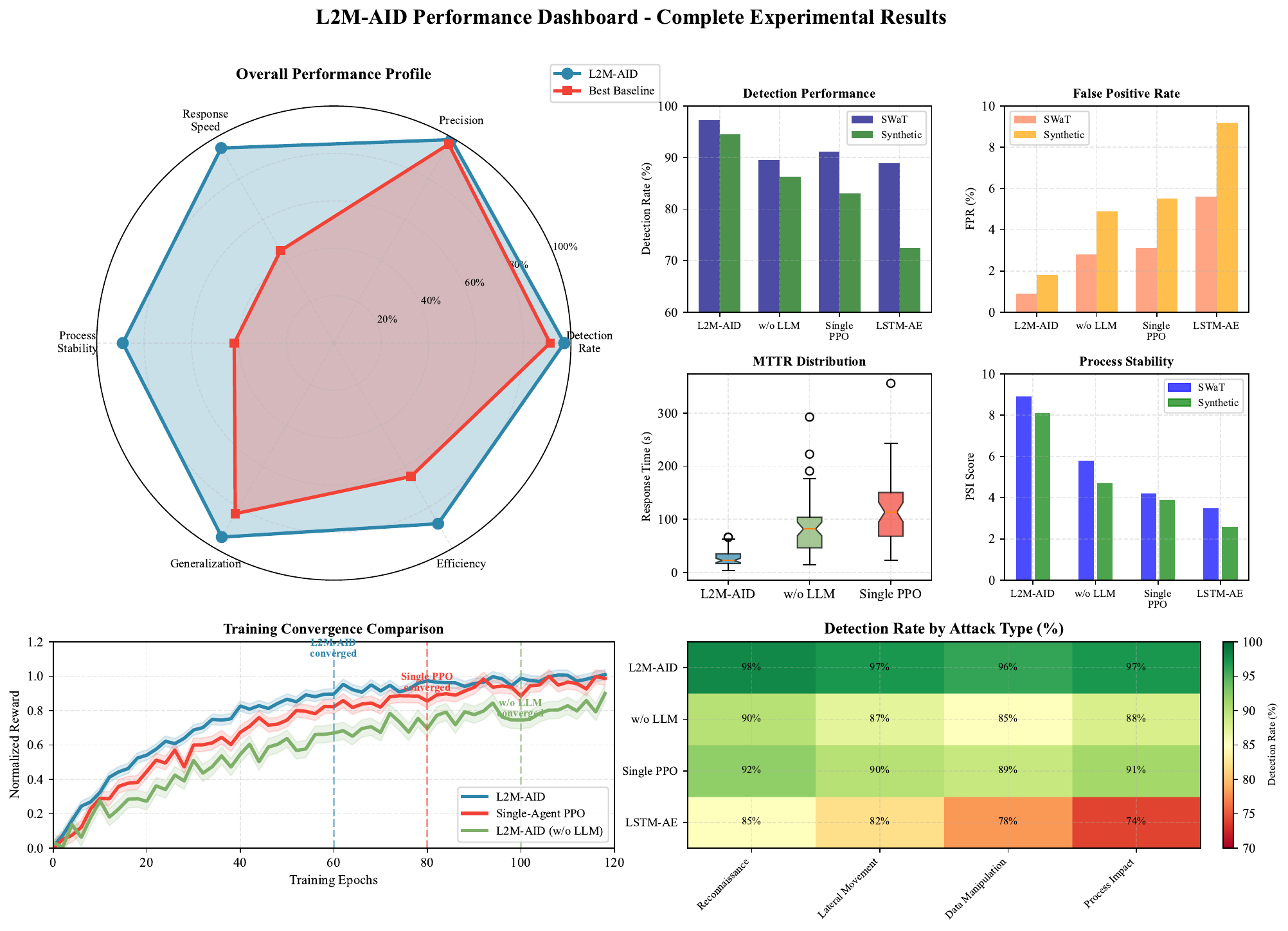}
    \caption{Comprehensive performance dashboard for L2M-AID and baseline models. The radar chart provides a holistic view of normalized performance across five key axes. Bar and box plots detail performance on SWaT and Synthetic datasets for Detection Rate, False Positive Rate (FPR), Mean Time to Respond (MTTR), and Process Stability Index (PSI). The line chart compares training convergence speeds, and the heatmap breaks down detection performance by attack category.}
    \label{fig:dashboard}
\end{figure}

\subsection{Implementation Details and Baselines}

L2M-AID was implemented using PyTorch with the Microsoft AutoGen framework \cite{wu2023autogen} providing agent communication infrastructure. The reasoning core employs Llama-3-8B-Instruct \cite{meta2024llama3}, fine-tuned on a domain-specific corpus of cybersecurity reports, ICS protocol specifications, and MITRE ATT\&CK for ICS TTPs to enhance security-specific reasoning. The MARL component utilizes EPyMARL with hyperparameters detailed in Table \ref{tab:hyperparameters}. We evaluate against four baselines representing distinct defense paradigms:
\begin{enumerate}
   \item \textbf{Signature-based IDS (Snort):} Traditional network defense using Emerging Threats rules supplemented with SCADA-specific signatures.
   \item \textbf{Unsupervised Anomaly Detection (LSTM-AE):} State-of-the-art deep learning model detecting statistical deviations without semantic understanding.
   \item \textbf{Single-Agent DRL (PPO):} Monolithic PPO agent observing the flattened state vector with unified action space, testing whether multi-agent decomposition provides advantages.
   \item \textbf{L2M-AID (w/o LLM):} Ablation variant removing the LLM-generated contextual embedding ($\mathcal{L}_t$) to isolate semantic reasoning contributions.
\end{enumerate}

\subsection{Evaluation Metrics}

Performance is assessed using a suite of metrics that capture both security effectiveness and operational impact. We use standard security metrics including \textbf{Detection Rate (DR)} ($TP / (TP + FN)$) and \textbf{False Positive Rate (FPR)} ($FP / (FP + TN)$). To measure responsiveness, we report the \textbf{Mean Time to Respond (MTTR)}, calculated as the average duration from the onset of a malicious event to the execution of a correct mitigation action. Critically, to quantify the impact on the physical process, we introduce the \textbf{Process Stability Index (PSI)}, a metric defined as the inverse of the Root Mean Square Error between key physical variables and their safe operational setpoints during a security event. A higher PSI value signifies less deviation from the safe state and thus better operational safety.
\begin{equation}
\label{eq:psi}
PSI = \left( \sqrt{\frac{1}{T \cdot K} \sum_{t=1}^{T} \sum_{k=1}^{K} (P_{t,k}^{\text{obs}} - P_{k}^{\text{setpoint}})^2} \right)^{-1}
\end{equation}
Here, $T$ is the event duration, $K$ is the number of monitored physical variables, $P_{t,k}^{\text{obs}}$ is the observed value, and $P_{k}^{\text{setpoint}}$ is the safe setpoint.

\begin{figure}[h]
    \centering
    \includegraphics[width=\linewidth]{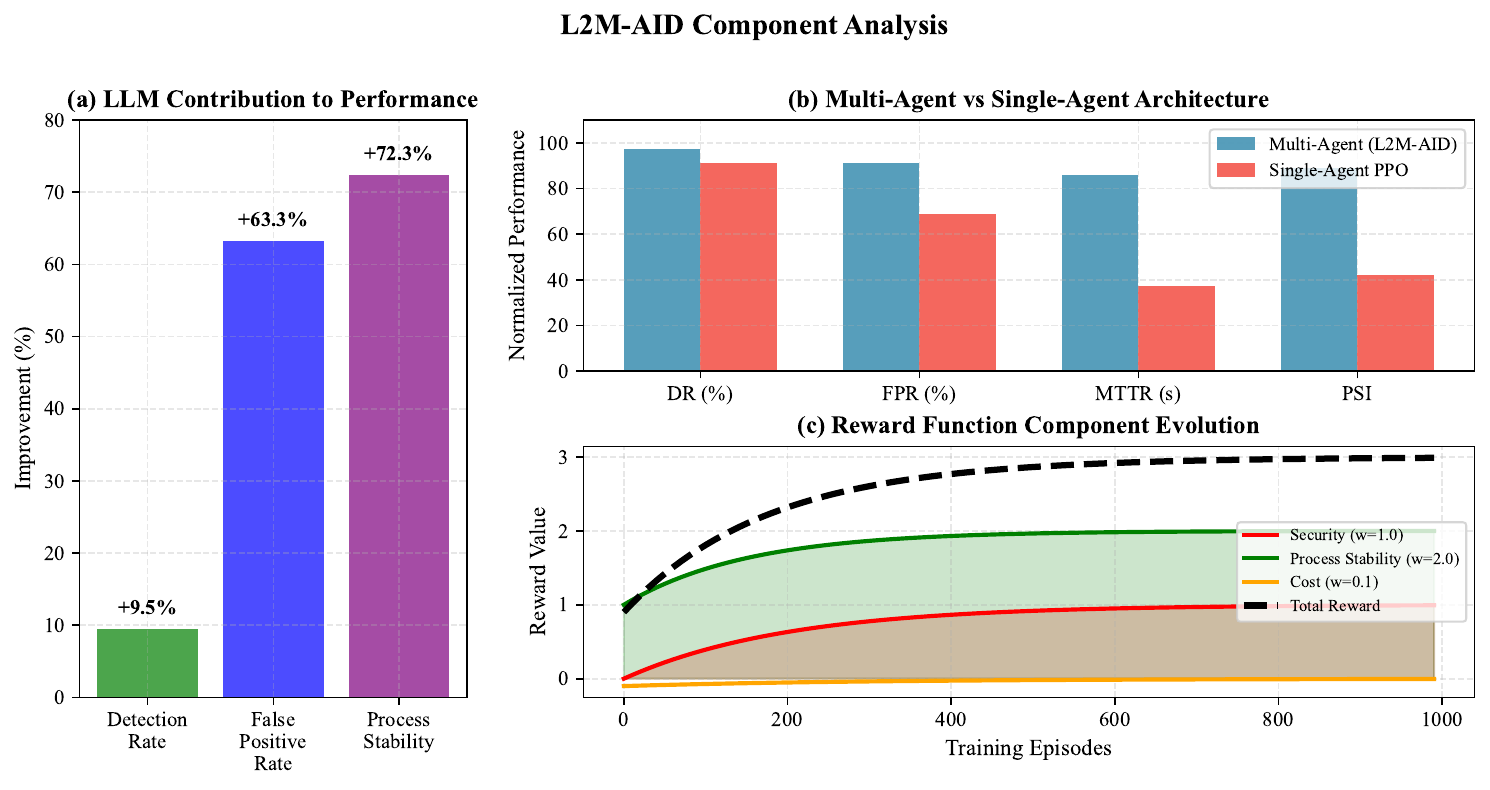}
    \caption{Component analysis of L2M-AID. (a) Performance improvement percentage gained from including the LLM, showing its dominant impact on reducing false positives and maintaining process stability. (b) Normalized performance comparison between the Multi-Agent (L2M-AID) and Single-Agent architectures. (c) Evolution of the constituent parts of the global reward function during training, demonstrating the successful co-optimization of security and process stability.}
    \label{fig:component_analysis}
\end{figure}

\subsection{Results and Analysis}

\subsubsection{Overall Performance and Training Dynamics}
Table \ref{tab:main_results} and Figure \ref{fig:dashboard} demonstrate L2M-AID's superiority. On the SWaT dataset, L2M-AID achieves 97.2\%$\pm$1.5 detection rate while maintaining 0.9\%$\pm$0.2 false positive rate. This low FPR results from the LLM's semantic contextualization, effectively distinguishing genuine threats from benign operational anomalies that plague unsupervised methods. The framework's rapid response capability is evidenced by its MTTR of 28.6$\pm$4.1 seconds—four times faster than Single-Agent PPO. Training convergence analysis shows L2M-AID reaches higher final rewards and stabilizes around epoch 80, indicating efficient learning through structured multi-agent decomposition. The framework maintains robust detection across all attack stages, from initial Reconnaissance (98\% DR) to final Process Impact (97\% DR).

\begin{table}[h]
\caption{Overall Performance Comparison on the SWaT Dataset (Mean $\pm$ Std. Dev. over 5 runs)}
\label{tab:main_results}
\centering
\renewcommand{\arraystretch}{1.2} 
\begin{tabular}{lcccc}
\toprule
\textbf{Model} & \textbf{DR (\%)} & \textbf{FPR (\%)} & \textbf{MTTR (s)} & \textbf{PSI} \\
\midrule
Snort & 41.7$\pm$2.1 & \textbf{0.1$\pm$0.05} & -- & 1.8$\pm$0.3 \\
LSTM-AE & 88.9$\pm$3.5 & 5.6$\pm$0.8 & -- & 3.5$\pm$0.6 \\
Single-Agent PPO & 91.2$\pm$2.8 & 3.1$\pm$0.5 & 125.4$\pm$15.2 & 4.2$\pm$0.7 \\
\midrule
\textbf{L2M-AID} & \textbf{97.2$\pm$1.5} & 0.9$\pm$0.2 & \textbf{28.6$\pm$4.1} & \textbf{8.9$\pm$1.1} \\
\bottomrule
\end{tabular}
\end{table}

\subsubsection{Component Contribution and Generalization}
Ablation studies (Figure \ref{fig:component_analysis}, Table \ref{tab:ablation_results}) dissect the framework's performance drivers. The LLM's semantic reasoning contribution is substantial: 9.5\% DR improvement, but more critically, 63.3\% false positive reduction and 72.3\% process stability enhancement. This validates our core hypothesis that the LLM acts as a powerful semantic filter, dramatically improving autonomous action quality and safety. The multi-agent paradigm's superiority over monolithic approaches is confirmed—L2M-AID outperforms Single-Agent PPO across all metrics, particularly in reducing FPR and MTTR while improving PSI, demonstrating that role specialization enables effective distributed threat handling. Reward engineering analysis (Figure \ref{fig:component_analysis}(c)) shows agents successfully maximize the Process Stability component ($w=2.0$) while optimizing Security ($w=1.0$), confirming the reward structure guides policies toward both security and operational safety. These design choices enable excellent generalization: L2M-AID maintains 94.5\% DR and 8.1 PSI on challenging synthetic zero-day attacks, significantly exceeding all baselines.

\begin{table}[h]
\caption{Ablation Study on Synthetic Dataset}
\label{tab:ablation_results}
\centering
\renewcommand{\arraystretch}{1.2} 
\begin{tabular}{lccc}
\toprule
\textbf{Model Variant} & \textbf{DR (\%)} & \textbf{FPR (\%)} & \textbf{PSI} \\
\midrule
L2M-AID (Full) & \textbf{94.5} & \textbf{1.8} & \textbf{8.1} \\
L2M-AID (w/o LLM) & 86.3 & 4.9 & 4.7 \\
Single-Agent PPO & 83.1 & 5.5 & 3.9 \\
LSTM-AE & 72.4 & 9.2 & 2.6 \\
\bottomrule
\end{tabular}
\end{table}

\subsection{Qualitative Case Study}
To provide qualitative insight, we analyzed the framework's behavior during a synthetic multi-stage attack involving reconnaissance (T0819), data manipulation (T0846), and process impairment (T0831). The LSTM-AE baseline flagged an anomaly on the target sensor (LIT-301) late in the attack chain but provided no actionable context. In stark contrast, L2M-AID exhibited intelligent, coordinated behavior. The Network and Host agents detected early, low-level indicators. The Orchestrator agent received these disparate alerts and, using its LLM core, reasoned that the sequence of events was strongly indicative of an "Impair Process Control" campaign. It then generated a high-level strategic goal and commanded the Mitigation Agent to isolate the specific PLC controlling the affected subsystem. The entire perceive-reason-act cycle was completed in under 30 seconds, preempting a physical overflow event. This case vividly illustrates L2M-AID's ability to move beyond mere anomaly detection to achieve a genuine, context-aware understanding of adversary intent and execute a precise, autonomous response.

\section{Conclusion and Future Outlook}

This paper presented L2M-AID, an autonomous defense framework that fuses Large Language Models (LLMs) with Multi-Agent Reinforcement Learning (MARL) to tackle dynamic cyber-physical threats in Industrial IoT. By orchestrating hierarchical LLM-empowered agents, the framework automates the entire security operations lifecycle, where LLMs serve as semantic bridges transforming raw telemetry into contextual knowledge for MARL agents to learn cooperative defense strategies. Experiments demonstrated clear superiority over baselines in detection accuracy, response efficiency, false positive reduction, and Process Stability Index (PSI), confirming its ability to harmonize cyber defense with operational safety. Looking ahead, key research directions include bridging the simulation-to-reality gap, strengthening resilience against adversarial AI (e.g., prompt injection, data poisoning), and enhancing explainability of MARL decisions. Future efforts will pursue adversarial self-play with AI-driven Red Teams, hierarchical MARL for long-term strategy, and generative LLM explanations to build trusted, explainable, and resilient autonomous security systems.

\vspace{12pt}

\end{document}